\documentclass{article}
\usepackage{spconf,amsmath,graphicx}
\usepackage{color}
\usepackage{multirow}
\usepackage{cite}
\usepackage[breaklinks=true,letterpaper=true,colorlinks,bookmarks=false,urlcolor=red,citecolor=cyan]{hyperref}
\usepackage{amsfonts,amssymb}
\usepackage{caption}
\usepackage{enumitem}
\setitemize{noitemsep,topsep=0pt,parsep=0pt,partopsep=0pt}

\usepackage{textcomp,subfigure,multirow,upgreek}
\usepackage{booktabs}
\usepackage{mdwlist}
\usepackage{mathrsfs,enumitem}

\usepackage{amssymb}

\usepackage{caption,color}

\usepackage{enumitem}
\setitemize{noitemsep,topsep=0pt,parsep=0pt,partopsep=0pt}

\title{PI-Trans: Parallel-ConvMLP and Implicit-Transformation Based GAN for Cross-View Image Translation}

\name{Bin Ren$^{1,2}$, Hao Tang$^{3}$, Yiming Wang$^{4}$, Xia Li$^{3}$, Wei Wang$^{5}$, Nicu Sebe$^{2}$\thanks{This work has been supported by the EU H2020 project AI4Media (No. 951911) and by the PRIN project PREVUE (Prot. 2017N2RK7K).}}
\address{
$^{1}$University of Pisa,
$^{2}$University of Trento,
$^{3}$ETH Zurich,
$^{4}$FBK,
$^{5}$Beijing Jiaotong University
}

\begin{document}
\maketitle

\begin{abstract}
For semantic-guided cross-view image translation, it is crucial to learn where to sample pixels from the source view image and where to reallocate them guided by the target view semantic map, especially when there is little overlap or drastic view difference between the source and target images. Hence, one not only needs to encode the long-range dependencies among pixels in both the source view image and target view semantic map but also needs to translate these learned dependencies. To this end, we propose a novel generative adversarial network, PI-Trans, which mainly consists of a novel Parallel-ConvMLP module and an Implicit Transformation module at multiple semantic levels. Extensive experimental results show that PI-Trans achieves the best qualitative and quantitative performance by a large margin compared to the state-of-the-art methods on two challenging datasets.
The source code is available at~\url{https://github.com/Amazingren/PI-Trans}.
\end{abstract}
\section{Introduction}
\label{sec:intro}
Semantic-guided cross-view image translation aims at generating images from a source view to a different target view given a target view semantic map as the guidance. 
In particular, we focus on the cases of translating from the aerial-view to the ground-view for photo-realistic urban scene synthesis, which can be beneficial for geo-localization~\cite{shi2019spatial,shi2020optimal,liu2019stochastic,zhu2021vigor,liu2019lending,hu2018cvm,cai2019ground} or civil engineering design with the semantic map either being extracted from another modality or being designed~\cite{toker2021coming,zhou2016view,regmi2018cross,turkoglu2019layer}.

\begin{figure}[!t]
	\centering
	\includegraphics[width=0.9\linewidth]{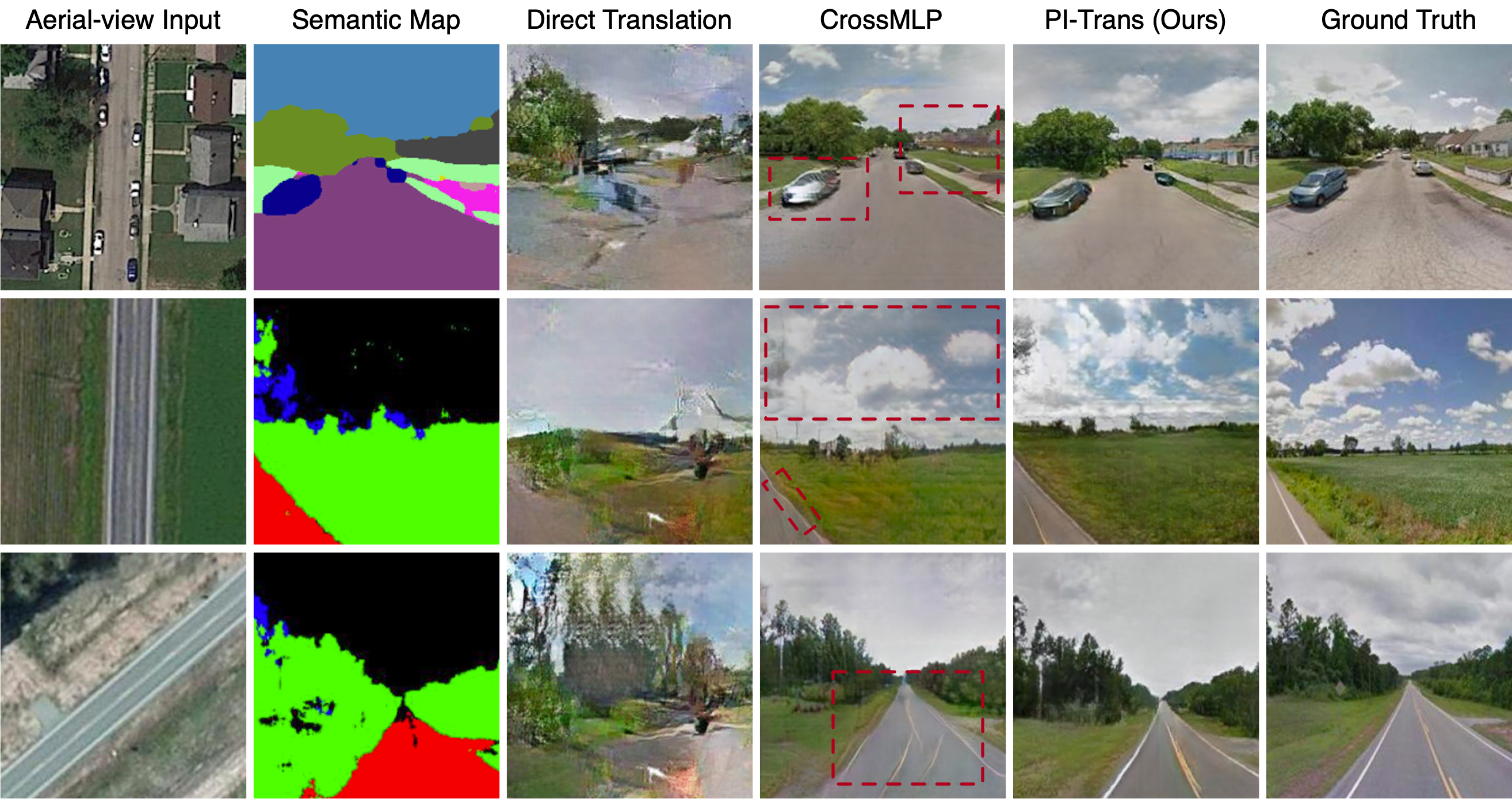}
	\caption{Examples of the cross-view translation with different methods. The aerial-view image (column 1) and semantic map (column 2) serve as the inputs. Direct translation (column 3) generates the target ground view without the guidance of the semantic map. CrossMLP~\protect\cite{ren2021cascaded} (column 4) is the most recent state-of-the-art method, yet its results still contain some obvious artifacts marked with dotted red boxes. Our PI-Trans (column 5) generates results that are most similar to the ground-truth images (column 6).}
	\label{fig:simpleShow}
	 \vspace{-0.4cm}
\end{figure}

However, translating images from two distinct views with little overlap is a challenging problem, as the area coverage, the appearances of objects, and their geometrical arrangement in the ground-view image can be extremely different from the aerial-view image (see the comparison between the 1st column and the 3rd column in Fig.~\ref{fig:simpleShow}). Early works usually adopt the convolutional neural networks (CNN) based encoder-decoder structure~\cite{zhou2016view,yang2015weakly} with the generative adversarial networks (GANs)~\cite{park2017transformation,trung2019learning}. However, such methodology suffers from satisfactory results when there exists little overlap between two views since the CNN-based methods struggle to establish the long-range relation due to the design nature of the convolutional kernels.

\begin{figure*}[!t] 
	\centering
	\includegraphics[width=0.7\linewidth]{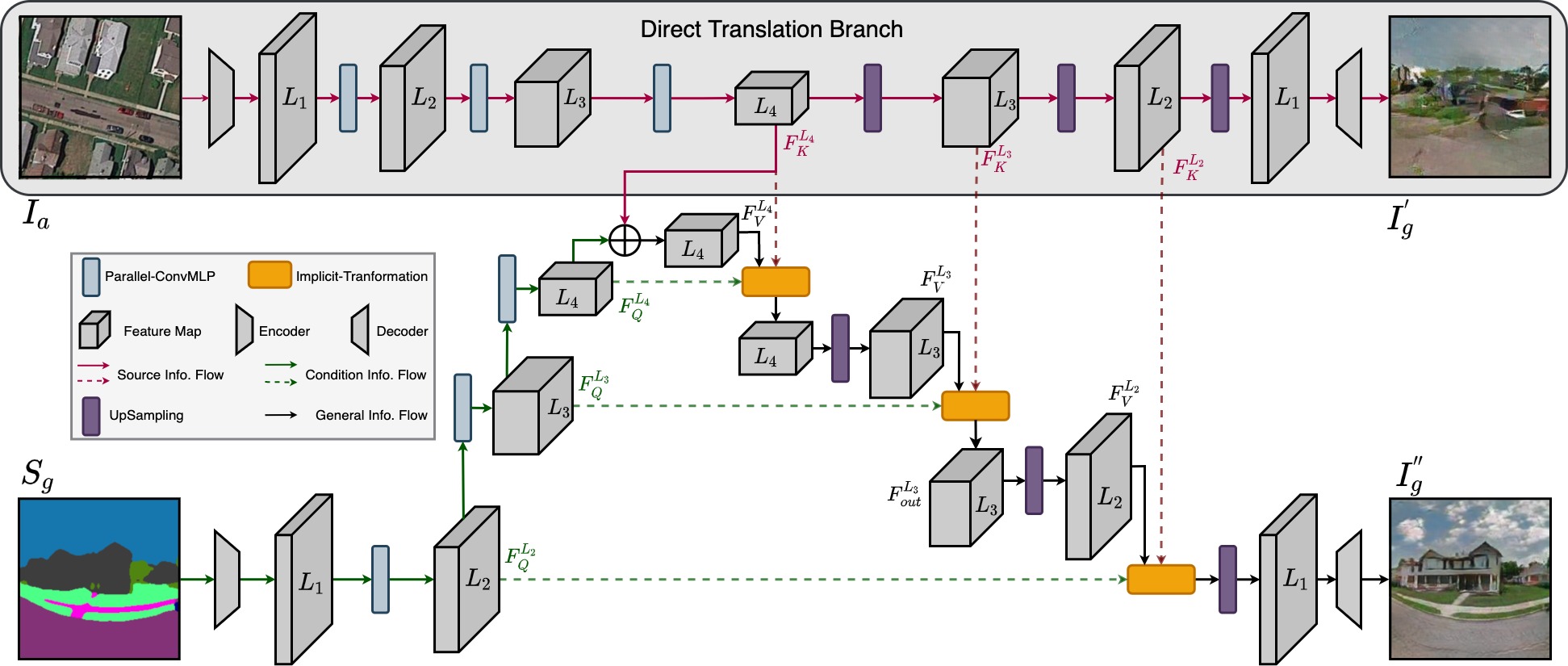}
	\caption{Architecture of our PI-Trans. The proposed parallel-ConvMLP module depicted in blue is assembled in two encoder branches for effectively modeling both the long-range relation and the fine spatial information, the proposed implicit-transformation module that depicted in yellow is used in two decoder branches for transforming the source-view information to target-view at multiple feature levels.
	$\oplus$ denotes the element-wise addition. $I_g^{'}$ is the output of the direct translation branch, $I_g^{''}$ is the final output of the proposed PI-Trans.}
	\label{fig:framework}
	\vspace{-0.4cm}
\end{figure*}

Usually, it's not easy to generate a photo-realistic image when just following the direct translation setting (the direct translation means that the generated ground-view image $I_g^{'}$ is directly generated from the aerial view image $I_a$ without the assistance of the semantic map $S_g$, see the direct translation branch in Fig.~\ref{fig:framework}). To improve the quality of the cross-view translation task, previous methods~\cite{regmi2018cross,tang2019multi,ren2021cascaded} took the target-view semantic map into consideration.

Though very insightful explorations had been performed, we find there are still some limitations that hinder the improvement of the quality of the generated images: (i) The pure CNN based methods (\emph{i.e.}, \cite{regmi2018cross,tang2019multi}) are difficult to establish the long-range relation due to its natural physical design of the convolutional kernels. (ii) The heavy fully-connected layers relied method~\cite{ren2021cascaded} is subject to be insufficient for modeling the fine spatial information. (iii) All these three state-of-the-art methods mentioned above missed utilizing the very crucial but the easiest to be ignored \emph{direct translation information} (the direct translation branch is shown in Fig.~\ref{fig:framework}).

To this end, we propose the Parallel-ConvMLP and Implicit-Transformation based GAN (PI-Trans), and its structure is shown in Fig.~\ref{fig:framework}, which is mainly built with two encoder branches and two decoder branches. In the encoder branches, we propose the novel parallel-ConvMLP module that can effectively manage both the latent long-range relation and the detailed spatial information with its unique spatial-channel MLP in a novel parallel manner instead of the sequential order like~\cite{ren2021cascaded,tolstikhin2021mlp}. Besides, the input feature is uniformly split into two chunks, and each is fed to its corresponding MLP. By doing so, we implicitly introduced channel shuffling, thus being beneficial for long-range reasoning. In the decoder branches, unlike previous methods that just take the combined feature from the source-view image and the target-view semantic map for recovering the final target-view image, we, for the first time, add the direct translation information (for providing reasonable color distribution) via the proposed implicit-transformation modules at multiple feature levels. 

To summarize, our contributions are listed as follows:
\begin{itemize}[leftmargin=*]
	\item We propose PI-Trans, which uses the transformation information that is directly learned from the source view to the target-view images to boost the performance.
 
    \item For modeling the long-range relation and the fine spatial information, we propose a novel parallel-ConvMLP module with an effective combination of CNN and MLP. Besides, an implicit-transformation module that conducts attention-based fusion at multiple feature levels is also proposed for a better translation result.

    \item Experimental results show that our method generates photo-realistic target-view ground images, scoring new state-of-the-art numerical results on two challenging datasets.
	
\end{itemize}

\section{The Proposed PI-Trans}
\vspace{-0.5em}
The architecture of our PI-Trans is shown in Fig.~\ref{fig:framework}, which consists of two encoder branches and two decoder branches. PI-Trans takes as input both the source-view aerial image $I_a{\in}{\mathbb{R}^{3 \times H \times W}}$ and the conditional target-view ground semantic map $S_g{\in}{\mathbb{R}^{3 \times H \times W}}$ at two encoder branches. First $I_a$ and $S_g$ are processed by two encoders to semantic $L_1$ level with dimension ($C_{L1}$, $H/2$, $W/2$), where $C$, $H$, and $W$ mean the channel number, height, width, respectively. We then use the proposed parallel-ConvMLP module (Sec.~\ref{sec:paramlp}) to further encode the $L_{1}$ level feature to $L_2$ (2$C_{L1}$, $H/4$, $W/4$), $L_3$ (4$C_{L1}$, $H/8$, $W/8$), and $L_4$ (8$C_{L1}$, $H / 16$, $W/16$) semantic levels. Unlike previous methods~\cite{ren2021cascaded,tang2019multi} which generate the final target-view ground image $I_g^{''}$ only based on the combined feature coming from $I_a$ and $S_g$ at $L_4$ level (\emph{i.e.}, $F_{V}^{L_{4}}$). We exploit another pathway, the direct translation branch, that directly produces a ground image $I_g^{'}$ at the target view from the source view, without interacting with the conditional semantic map. Then we use this direct transformation information accompanied with the semantic feature at the lower target pathway via our proposed implicit-transformation (Sec.~\ref{sec:impletrans}) at 3 semantic levels ($L_2$, $L_3$, and $L_4$). 

\subsection{Parallel-ConvMLP Module}
\label{sec:paramlp}
\begin{figure}[!t] 
	\centering
	\includegraphics[width=1.0\linewidth]{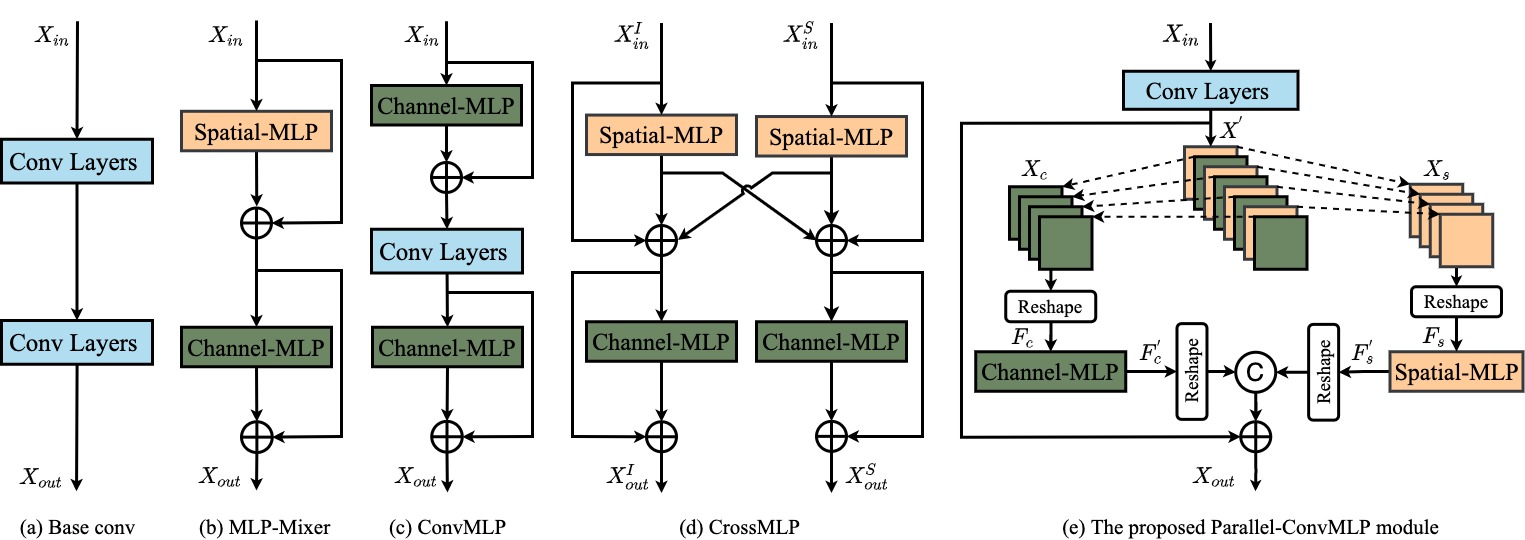}
	\caption{Illustrations of four comparison methods \emph{i.e.}, (a) Basic convolutional layers, (b) MLP-Mixer \protect\cite{tolstikhin2021mlp}, (c) ConvMLP \protect\cite{li2021convmlp}, (d) CrossMLP module~\protect\cite{ren2021cascaded} and (e) our parallel-ConvMLP. The symbols $\oplus$ and \textcircled{c} denote the element-wise addition and channel-wise concatenation.}
	\label{fig:paraMLP}
	\vspace{-0.4cm}
\end{figure}

Given one pixel, it's extremely significant for the cross-view image translation task to understand which other pixels are related to it, or which object it belongs to. However, CNN kernels (Fig.~\ref{fig:paraMLP}(a)) are not good at modeling the long-range relation because of the locality of the fixed kernels. Therefore, the MLP-based methods MLP-Mixer~\cite{tolstikhin2021mlp} (Fig.~\ref{fig:paraMLP}(b)), ConvMLP~\cite{li2021convmlp} (Fig.~\ref{fig:paraMLP}(c)), and CrossMLP~\cite{ren2021cascaded} (Fig.~\ref{fig:paraMLP}(d)) were proposed to ease this problem. However, both MLP-Mixer and CrossMLP are computationally heavy and are not subject to modeling the fine spatial patterns. ConvMLP~\cite{li2021convmlp} tackles this problem by combining a depth-wise convolution between two channel-wise MLPs. Yet, its performance is still unsatisfactory for the cross-view translation task (See Sec.~\ref{sec:exper}).

To explore the balance between the long-range relation and the spatial pattern modeling, we propose a parallel-ConvMLP module, which is shown in Fig.~\ref{fig:paraMLP}(e). Given an input feature $X_{in}$ in dimension ($c, h, w$), it firstly goes through two convolutional layers:
\begin{equation}
\begin{aligned}
X^{'} = {\rm Conv}({\rm downConv}(X_{in})),
\end{aligned}
\label{eqn:1}
\end{equation} 
The first ($\rm downConv(\cdot)$) is a strided convolutional layer while the second ($\rm Conv(\cdot)$) is a normal one. $X^{'}$ is the output in dimension ($2c, h/2, w/2$). This operation ensures the latent appearance and the local spatial pattern can be well established. Different from previous methods that simultaneously model the long-range relation and the spatial information in a sequential way~\cite{ren2021cascaded,tolstikhin2021mlp}, we decouple the problem by dividing $X^{'}$ along channel dimension into two parts based on the parity of the channel index: 
\begin{equation}
\begin{aligned}
X_{c} = X^{'}_{2i-1},~~X_{s} = X^{'}_{2i},~~for~~i=1, 2, 3,\cdots,2c.
\end{aligned}
\label{eqn:2}
\end{equation}

After that, we first flatten $X_c$ and $X_s$ from ($c, h/2, w/2$) to ($n, c$) and ($c, n$), forming $F_c$ and $F_s$ ($n{=} w/2 {\times} h/2$). Then two fully-connected MLPs (channel-wise and spatial-wise) are applied to $F_c$ and $F_s$. And each MLP block contains two fully-connected layers and a nonlinear GELU~\cite{hendrycks2016gaussian} activation function. These operations are formulated as follows:
\begin{equation}
\begin{aligned}
F_c^{'} &= {\rm \boldsymbol{W_2^{c}}} \sigma({\rm \boldsymbol{W_1^{c}}} (F_c)_{*, i}),~~for~~i = 1, 2, 3, \cdots, c,\\
F_s^{'} &= {\rm \boldsymbol{W_2^{s}}} \sigma({\rm \boldsymbol{W_1^{s}}} (F_s)_{*, j}),~~for~~j = 1, 2, 3, \cdots, n,
\end{aligned}
\label{eqn:3}
\end{equation}
where ${\rm \boldsymbol{W_1^{c}}}$ and ${\rm \boldsymbol{W_2^{c}}}$ denote the learnable weights for the channel-wise MLP, while ${\rm \boldsymbol{W_1^{s}}}$ and ${\rm \boldsymbol{W_2^{s}}}$ are for the spatial-wise MLP. $\sigma(\cdot)$ means the GELU activation function. Then $F_c^{'}$ and $F_s^{'}$ are reshaped back to ($c, h, w$) and concatenated together. Finally, a skip-connection is used to add the concatenated feature to $X^{'}$:
\begin{equation}
\begin{aligned}
X_{out} = X^{'} + {\rm Cat}({\rm Reshape}(F_c^{'}), {\rm Reshape}(F_s^{'})),
\end{aligned}
\label{eqn:4}
\end{equation}
where $\rm Cat(\cdot)$ and $\rm Reshape(\cdot)$ denote the concatenate and reshape operations.

\subsection{Implicit Transformation Module}
\label{sec:impletrans}
Since the performance for the direct translation ($I_a \to I_g^{'}$) is in a bad condition (see $I_g^{'}$ in Fig.~\ref{fig:framework} or the third column in Fig.~\ref{fig:simpleShow}). Hence, previous methods SelectionGAN~\cite{tang2019multi} and CrossMLP~\cite{ren2021cascaded} ignored the direct translation branch and just used the combined feature $F_{V}^{L_{4}}$. Instead, we explore how this kind of latent information within the direct transformation branch may impact the overall generation performance since it provides a reasonable color distribution though bad in spatial structure. Therefor, we propose the implicit transformation module (the yellow blocks in Fig.~\ref{fig:framework}), which conducts the fusion at multiple semantic levels.

There are three implicit transformation modules at $L_4$, $L_3$, and $L_2$ levels, each of them takes as input three kinds of information, \emph{i.e.}, the ground semantic map feature $F_Q$, the directly translated feature $F_K$, and an extra input feature $F_V$. In Fig.~\ref{fig:framework}, we visualize these three kinds of information flows in green, red, and black colors, respectively. The main idea behind the proposed implicit transformation module is to enable the transformed feature $F_K$ to provide useful latent appearance or color information for generation. More specifically, we exploit the semantic map feature $F_Q$ and the directly transformed feature $F_K$ to construct an attention map with a softmax function. This operation uses the target-view semantic feature $F_Q$ to select the most important information in the directly transformed feature $F_K$, which can also be seen as a learned latent transformation pattern. This attention map is then used to activate the most relevant feature in $F_V$, achieving the feature-level implicit transformation guided by the attention map. Finally, we exploit a skip-connection to maintain the result in the last module. And we take the $L_3$ level module for a detailed description. Given three inputs $F_{Q}^{L_{3}}$, $F_{K}^{L_{3}}$, and $F_{V}^{L_{3}}$, we first reshape them from ($b, c, h, w$) to ($b, c/4, n$), ($b, c/4, n$), and ($b, c, n$). Here $n {=} h {\times} w$. Then we adopt the residual attention mechanism to learn the fused feature:
\begin{equation}
\begin{aligned}
F_{out}^{L_{3}} = F_{V}^{L_{3}} + {\rm softmax} (F_{Q}^{L_{3}} (F_{K}^{L_{3}})^{T}) F_{V}^{L_{3}},
\end{aligned}
\label{eqn:5}
\end{equation}
where $F_{out}^{L_{3}}$ denotes the output. It serves as the value feature for the implicit translation module at the next semantic level. 

\subsection{Discriminator and Optimization Objective}
\noindent\textbf{Discriminator.} Following~\cite{regmi2018cross} and \cite{tang2019multi}, the discriminator in the direct transformation branch takes the real image $I_a$ and the generated image $I_g^{'}$ or the ground-truth image $I_g$ as input. While for the lower decoder branch, it accepts the real image $I_a$ and the generated image $I_g^{''}$ or the real image $I_g$ as input.

\noindent\textbf{Optimization Objective.} The full optimization objective is:
\begin{equation}
\begin{aligned}
\min_{\{G\}} \max_{\{D\}} \mathcal{L} =  \lambda_{1}\mathcal{L}_{1} +  \lambda_{cGAN}\mathcal{L}_{cGAN} + \mathcal{L}_{tv} + \lambda_{per}\mathcal{L}_{per},
\end{aligned}
\label{eqn:6}
\end{equation}
where $\mathcal L_1$ denotes the pixel-level loss, $\mathcal{L}_{cGAN}$ denotes the adversarial loss, which is used for distinguishing the synthesized images pairs ($I_a, I_g^{''}$) from the real image pairs ($I_a, I_g$). $\mathcal{L}_{tv}$ is the total variation regularization~\cite{johnson2016perceptual} and $\mathcal{L}_{per}$ is the perception loss which is commonly used for the generative tasks~\cite{johnson2016perceptual,siarohin2018deformable} to make the produced images look more natural and smooth. We set $\lambda_{1}$, $\lambda_{cGAN}$, and $\lambda_{per}$ to 100, 5, and 50, respectively.

\section{Experiments}
\label{sec:exper}
\noindent\textbf{Datasets and Evaluation Metrics.}
We conduct experiments on two challenging public datasets, \emph{i.e.}, Dayton~\cite{vo2016localizing}, and CVUSA\cite{workman2015wide}, following the similar settings as in \cite{regmi2018cross,tang2019multi}. 
We follow a similar evaluation protocol as in \cite{regmi2018cross,tang2019multi} to measure the quality of the generated ground images.

\noindent\textbf{Implementation Details.}
\textit{Encoders} are the same for both $I_a$ and $S_g$, and each consists of one two-strided down-sampling convolutional layer (followed by batch normalization and ReLU activation function) and two one-strided ones (filter numbers: 16, 32, 32) to generate the $L_1$ level features. Given an input in dimension ($b, 3, 256, 256$), the output is in dimension ($b, 32, 128, 128$), where $b$ denotes the batch size. 
Each \textit{Up-sampling block }(depicted in purple in Fig.~\ref{fig:framework} are used in the decoder branches) consists of one nearest up-sampling layer, two one-strided convolutional layers (kernel size: 3, 3). \textit{Decoder} consists of three one-strided convolutional layers (kernel size: 3, 3, and padding: 1, 1) and 1 Tanh activation function.

\noindent\textbf{Training Settings.} Following~\cite{regmi2018cross} and \cite{tang2019multi}, Adam~\cite{kingma2014adam} is used as the solver with momentum terms $\beta_1 = 0.5$ and $\beta_2 = 0.999$. The initial learning rate for is set to $0.0002$. In addition, we train the proposed PI-Trans 35 epochs on the Dayton dataset and 30 epochs on the CVUSA dataset in an end-to-end manner. The code is implemented in PyTorch~\cite{paszke2019pytorch}.

\begin{table}[!t] \small
	\centering
	\caption{Quantitative results on the Dayton dataset.}
	\resizebox{1.0\linewidth}{!}{%
		\begin{tabular}{cccccccccccc} \toprule
			\multirow{2}{*}{Method}  & \multicolumn{4}{c}{Accuracy (\%) $\uparrow$}& \multicolumn{3}{c}{Inception Score $\uparrow$} &  \multirow{2}{*}{KL $\downarrow$} & \multirow{2}{*}{LPIPS $\downarrow$} \\ \cmidrule(lr){2-5} \cmidrule(lr){6-8} 
			& \multicolumn{2}{c}{Top-1} & \multicolumn{2}{c}{Top-5} & all & Top-1 & Top-5  \\ \hline
			Pix2pix~\cite{isola2017image} &6.80$^\ast$ & 9.15$^\ast$& 23.55$^\ast$ &27.00$^\ast$ & 2.8515$^\ast$ &1.9342$^\ast$ &2.9083$^\ast$ & 38.26 $\pm$ 1.88$^\ast$ & -
			\\
			X-Fork \cite{regmi2018cross} &30.00$^\ast$ &48.68$^\ast$ & 61.57$^\ast$ & 78.84$^\ast$ &3.0720$^\ast$ &2.2402$^\ast$ &3.0932$^\ast$  & 6.00 $\pm$ 1.28$^\ast$ & -
			\\
			X-Seq \cite{regmi2018cross} &30.16$^\ast$ &49.85$^\ast$& 62.59$^\ast$&80.70 &2.7384$^\ast$ & 2.1304$^\ast$ &2.7674$^\ast$ & 15.93 $\pm$ 1.32$^\ast$ & - 
			\\
			SelectionGAN~\cite{tang2019multi}
			& 42.11 & 68.12 & 77.74 & 92.89 & 3.0613 & 2.2707 & 3.1336 & 2.7406 $\pm$ 0.8613 & 0.7961
			\\
            CrossMLP~\cite{ren2021cascaded} & 47.65 & 78.59 & 80.04 & 94.64 & 3.3466 & 2.2941 & 3.3783 & 2.3301 $\pm$ 0.8014 & 0.3750
			\\
			PI-Trans (Ours) & \textbf{49.48} & \textbf{79.98} & \textbf{82.45} & \textbf{96.52} & \textbf{3.6713} & \textbf{2.4918} & \textbf{3.6824} & \textbf{2.0790} $\pm$\textbf{ 0.6509} & \textbf{0.3656}
			\\
            \hline
            Real Data & - & -&-&-&3.8326&2.5781&3.9163&-&-\\
			\bottomrule		
	\end{tabular}}
	\label{tab:dayton}
	\vspace{-0.4cm}
\end{table}

\begin{table}[!t] \small
	\centering
	\caption{Quantitative results on the CVUSA dataset.}
	\resizebox{1.0\linewidth}{!}{%
		\begin{tabular}{cccccccccccc} \toprule
			\multirow{2}{*}{Method}  & \multicolumn{4}{c}{Accuracy (\%) $\uparrow$}& \multicolumn{3}{c}{Inception Score $\uparrow$} &  \multirow{2}{*}{KL $\downarrow$} & \multirow{2}{*}{LPIPS $\downarrow$} \\ \cmidrule(lr){2-5} \cmidrule(lr){6-8} 
			& \multicolumn{2}{c}{Top-1} & \multicolumn{2}{c}{Top-5} & all & Top-1 & Top-5  \\ \hline
			Pix2pix~\cite{isola2017image} &7.33$^\ast$ & 9.25$^\ast$& 25.81$^\ast$ &32.67$^\ast$ & 3.2771$^\ast$ &2.2219$^\ast$ &3.4312$^\ast$ & 59.81 $\pm$ 2.12$^\ast$ & -
			\\
			X-Fork \cite{regmi2018cross} &20.58$^\ast$ &31.24$^\ast$ & 50.51$^\ast$ & 63.66$^\ast$ &3.4432$^\ast$ &2.5447$^\ast$ &3.5567$^\ast$  & 11.71 $\pm$ 1.55$^\ast$ & -
			\\
			X-Seq \cite{regmi2018cross} &15.98$^\ast$ &24.14$^\ast$& 42.91$^\ast$&54.41 &3.8151$^\ast$ & 2.6738$^\ast$ &4.0077$^\ast$ & 15.52 $\pm$ 1.73$^\ast$ & - 
			\\
			SelectionGAN~\cite{tang2019multi}
			& 41.52 & 65.61 & 74.32 & 89.66 & 3.8074 & 2.7181 & 3.9197 & 2.9603 $\pm$ 0.9714 & 0.4122
			\\
			CrossMLP~\cite{ren2021cascaded} & 44.96 & 69.96 & 76.98 & 91.91 & 3.8392 & 2.8035 & 3.9757 & 2.6903 $\pm$ 0.9432 & 0.3974
			\\
			PI-Trans (Ours) & \textbf{47.87} & \textbf{74.57} & \textbf{80.36} & \textbf{94.68} & \textbf{4.1701} & \textbf{2.9878} & \textbf{4.2071} & \textbf{2.2363} $\pm$ \textbf{0.7759}& \textbf{0.3784}
			\\
            \hline
            Real Data & - & -&-&-&4.8749&3.3053&4.9938&-&-\\
			\bottomrule		
	\end{tabular}}
	\label{tab:CVUSA}
	\vspace{-0.4cm}
\end{table}

\begin{figure}[!t] \small
	\centering
	\includegraphics[width=1.0\linewidth]{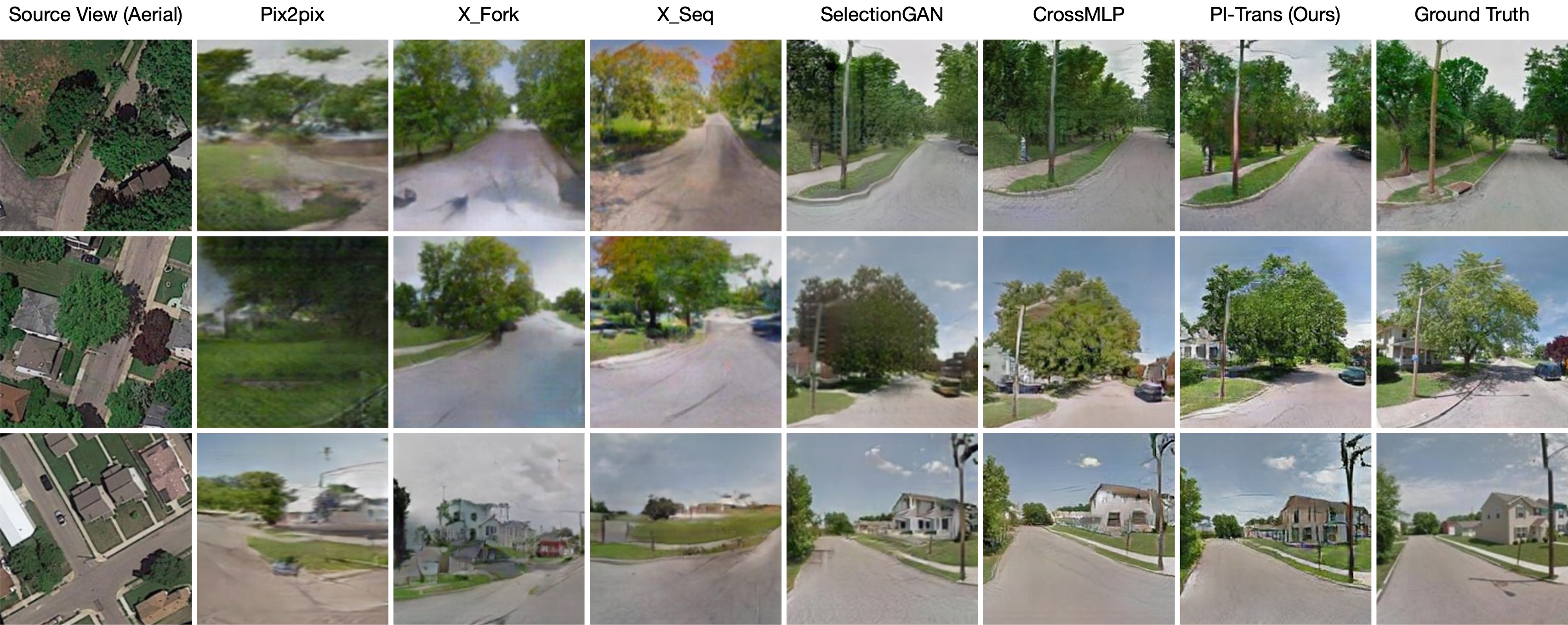}
	\caption{Qualitative results of different methods on Dayton.
	}
	\label{fig:sota_dayton}
    \vspace{-0.4cm}
\end{figure}

\begin{figure}[!t] \small
	\centering
	\includegraphics[width=1.0\linewidth]{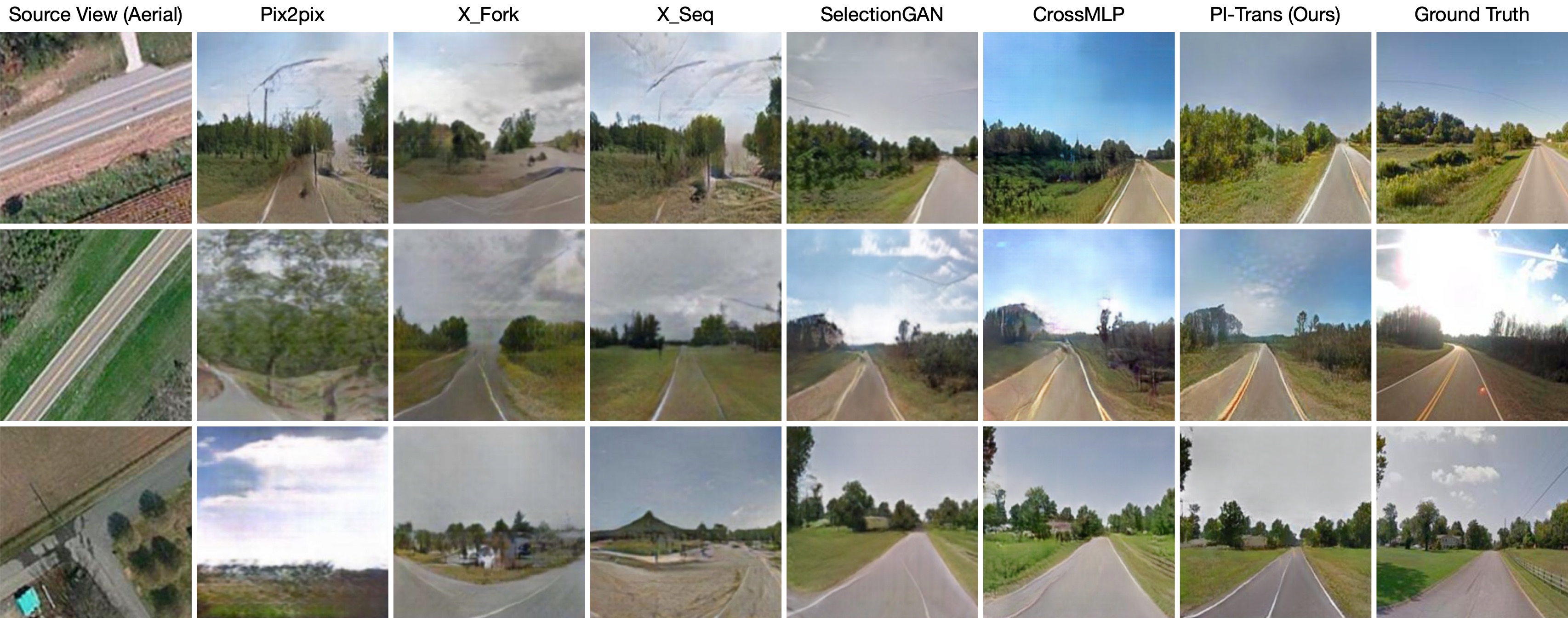}
	\caption{Qualitative results of different methods on CVUSA.
	}
	\label{fig:sota_cvusa}
	\vspace{-0.4cm}
\end{figure}

\noindent\textbf{Experimental Results.}
PI-Trans is compared with five state-of-the-art methods, including Pix2pix~\cite{isola2017image}, X-Fork~\cite{regmi2018cross}, X-Seq~\cite{regmi2018cross}, SelectionGAN~\cite{tang2019multi}, and CrossMLP~\cite{ren2021cascaded}.
Table~\ref{tab:dayton} and Table~\ref{tab:CVUSA} show the numerical results on Dayton and CVUSA. 
PI-Trans achieves the best performance on all the metrics with a significant improvement.
The qualitative results are shown in Fig.~\ref{fig:sota_dayton} and Fig.~\ref{fig:sota_cvusa} for Dayton and CVUSA, respectively. The target-view ground images generated by our PI-Trans are more natural and sharper.

\noindent\textbf{Ablation Study.}
To validate the effectiveness of each proposed module, we select 1/3 samples randomly from the entire Dayton dataset, forming the Dayton-Ablation dataset. 
Six baselines \emph{i.e.}, A (Basic Conv), B (MLP-Mixer), C (ConvMLP), D (CrossMLP), E (Parallel-ConvMLP), and F (PI-Trans) are proposed. For baseline A, we only use the convolutional layers, and without the implicit transformation modules. The output directly comes out from a combination of the features between source-view image features and the semantic map features. Baseline B, C, D, and E follow a similar setting as baseline A, with the only difference in the encoder branches. F is our full model. The results shown in Table~\ref{tab:ablation} demonstrate that both our Parallel-ConvMLP (E) and implicit transformation (F) can boosts the quality of the generated ground image by a large margin.

\begin{table}[!t] \small
	\centering
	\caption{Ablation results on Dayton-Ablation.}
	\resizebox{1.0\linewidth}{!}{%
		\begin{tabular}{lccccccccccc} \toprule
			\multirow{2}{*}{Method}  & \multicolumn{4}{c}{Accuracy (\%) $\uparrow$}& \multicolumn{3}{c}{Inception Score $\uparrow$} &  \multirow{2}{*}{KL $\downarrow$} & \multirow{2}{*}{LPIPS $\downarrow$} \\ \cmidrule(lr){2-5} \cmidrule(lr){6-8} 
			& \multicolumn{2}{c}{Top-1} & \multicolumn{2}{c}{Top-5} & all & Top-1 & Top-5  \\ \hline
			A: Basic Conv &43.61 & 71.85 &76.56 &91.55 & 2.8132 &2.1399 &2.8036 & 2.9343 $\pm$ 0.9248 & 0.4392
			\\
			B: MLP-Mixer &44.36 &72.08 & 77.57 & 92.44 &3.3207 &2.2852 &3.3282  & 2.5810 $\pm$ 0.8489 & 0.3926
			\\
			C: ConvMLP &42.75 &69.35& 76.49&89.82 &3.2442 & 2.2229 &3.2580 & 2.8470 $\pm$ 0.8619 & 0.4093 
			\\
			D: CrossMLP
			& 43.41 & 70.73 & 76.84 & 91.77 & 3.3241 & 2.2580 & 3.3340 & 2.7250 $\pm$ 0.8703 & 0.3987
			\\
			E: P-ConvMLP & 45.63 & 73.04 & 80.02 & 92.62 & 3.2573 & 2.2742 & 3.2957 & 2.4877 $\pm$ 0.8014 & 0.3917
			\\
			F: PI-Trans & \textbf{49.90} & \textbf{79.27} & \textbf{82.83} & \textbf{95.20} & \textbf{3.3975} & \textbf{2.3461} & \textbf{3.4111} & \textbf{2.0856} $\pm$ \textbf{0.7045} & \textbf{0.3828}
			\\
            \hline
            Real Data & - & -&-&-&3.8307&2.5749&3.9159&-&-\\
			\bottomrule		
	\end{tabular}}
	\label{tab:ablation}
	\vspace{-0.4cm}
\end{table}

\section{Conclusion}
\label{sec:conclusion}
We propose a new PI-Trans for generating realistic cross-view images. Significantly, the parallel-ConvMLP module is designed for balancing the relationship between latent long-range dependency modeling and spatial information maintenance. The implicit transformation module we designed at multiple semantic levels in this paper carefully takes care of not only the target-view semantic map feature and the source-view aerial image feature but also the direct translation information that is usually ignored by previous methods. We validate the effectiveness and advances of our proposed modules over existing methods.

\clearpage

\bibliographystyle{IEEEbib}
\bibliography{strings,refs}

\end{document}